\documentclass[twocolumn,letterpaper]{IEEEAerospaceCLS}  

\usepackage[]{graphicx}    
\usepackage[colorlinks=true, urlcolor=blue, linkcolor=black, citecolor=black]{hyperref} 
\newcommand{\ignore}[1]{}  
\usepackage{tabularx}
\usepackage{graphicx}
\usepackage{array}
\usepackage{svg}
\usepackage{tcolorbox}
\usepackage{threeparttable}
\usepackage{caption}
\usepackage{xcolor}
\usepackage{colortbl}
\usepackage{pifont}
\newcommand{\cmark}{\ding{51}}%
\newcommand{\xmark}{\ding{55}}%
\usepackage{booktabs}
\newcommand{\tabitem}{~~\llap{\textbullet}~~}
\usepackage{makecell}
\usepackage{tikz}
\usetikzlibrary{calc}

\begin{document}

\providecommand{\cmtt}[1]{{\fontfamily{cmtt}\selectfont #1}}

\title{The S3LI Vulcano Dataset: A Dataset for Multi-Modal SLAM in Unstructured Planetary Environments}

\author{%
Riccardo Giubilato\\ 
German Aerospace Center (DLR)\\
Institute of Robotics and Mechatronics \\
riccardo.giubilato@dlr.de 
\and 
Marcus Gerhard Müller\\ 
German Aerospace Center (DLR)\\
Institute of Robotics and Mechatronics \\
marcus.mueller@dlr.de 
\and 
Marco Sewtz\\ 
German Aerospace Center (DLR)\\
Institute of Robotics and Mechatronics \\
marco.sewtz@dlr.de 
\and 
Laura Alejandra Encinar Gonzalez \\
German Aerospace Center (DLR)\\
Institute of Robotics and Mechatronics \\
KTH Royal Institute of Technology \\
laura.encinargonzalez@dlr.de \\
\and 
John Folkesson \\
KTH Royal Institute of Technology \\
Deptartment of Intelligent Systems \\
johnf@kth.se \\
\and
Rudolph Triebel\\ 
German Aerospace Center (DLR)\\
Institute of Robotics and Mechatronics \\
rudolph.triebel@dlr.de 
\thanks{\footnotesize 979-8-3315-7360-7/26/$\$31.00$ \copyright2026 IEEE}              
}

\maketitle

\thispagestyle{plain}
\pagestyle{plain}

\begin{abstract}
\textbf{We release the S3LI Vulcano dataset, a multi-modal dataset towards development and benchmarking of Simultaneous Localization and Mapping (SLAM) and place recognition algorithms that rely on visual and LiDAR modalities. Several sequences are recorded on the volcanic island of Vulcano, from the Aeolian Islands in Sicily, Italy. The sequences provide users with data from a variety of environments, textures and terrains, including basaltic or iron-rich rocks, geological formations from old lava channels, as well as dry vegetation and water. The data (\url{rmc.dlr.de/s3li_dataset}) is accompanied by an open source toolkit (\url{github.com/DLR-RM/s3li-toolkit}) providing tools for generating ground truth poses as well as preparation of labelled samples for place recognition tasks.} 
\end{abstract}

\tableofcontents

\section{Introduction}

\begin{figure}
    \centering
    \includegraphics[width=\linewidth]{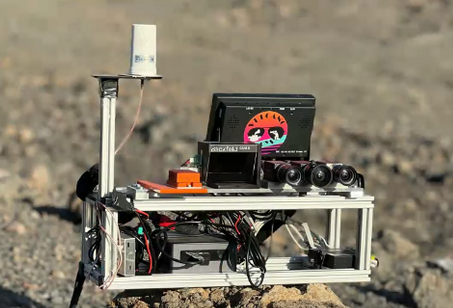}
    \caption{Impression of the \textbf{S3LI} (RGB \textbf{S}tereo, \textbf{S}olid-\textbf{S}tate \textbf{L}iDAR, \textbf{I}nertial) sensor setup, captured on the \textit{Gran Cratere della Fossa}, the active center of Vulcano, vulcan from the homonymous island from the Aeolian Islands, Sicily. On the setup it is visible the white GNSS antenna, computer unit from an Intel NUC and touchscreen to operate the sensor setup.}
    \label{fig:s3li_setup}
\end{figure}

Simultaneous Localization and Mapping (SLAM) components are necessary for mobile systems to achieve high-level autonomy, as they provide the ability to perform localization on a simultaneously growing map from an entirely unknown environment. While having achieved in the past year a fairly elevated level of maturity for traditional artificial environments (e.g., cityscapes or indoor environments), methods that achieve high performances in fully unstructured settings, such as for planetary-like scenes, belong to an open and active area of research \cite{giubilato2022gpgm,geromichalos2020slam,yin2025general}. 

Development and testing of SLAM algorithms in challenging outdoor environments is difficult due to these places of interest being often hardly reachable (e.g., volcanoes or deserts), and a solution to this is to rely on publicly available data. The majority of data available to challenge and evaluate the accuracy of multi-modal SLAM algorithms, however, often consider environments that are affected by anthropogenic factors. The introduction of human-made elements, both in urban but also natural contexts, often introduce perceptual and structural regularities that bias the performance of state estimation and mapping. When recording data from moving platforms, such as in the case of autonomous driving, the repetitiveness of the driving direction impose significant limitations to the variety of viewpoints to expect, significantly helping place recognition algorithms in re-detecting previously visited scenes to establish "Loop Closures" within SLAM. In planetary exploration settings, however, mobile systems operate in highly unstructured natural environments, which open large possibilities to traverse environments, posing significant challenges to the aforementioned tasks. Natural environments are characterized by extreme perceptual aliasing and abundance of ambiguous features, which, exacerbated by unconstrained traversing opportunities, pose significant limitations to SLAM algorithms to provide an accurate pose to the observer with respect to the origin of their map representation.

\begin{table*}[tp]
\caption{Evaluation of selected related works. Here is only considered data related to mobile robots, or hand-held platforms, excluding the case of road-bound vehicles.}
\begin{threeparttable}
\renewcommand{\arraystretch}{1.2}
\centering
    \setlength{\tabcolsep}{7.25pt}
\begin{tabular}{|c|cc|ccccc|}
 \toprule
 \hline
 \textbf{Name} & \textbf{Platform} & \textbf{Env.} & Mono & Stereo & IMU & Lidar & GT \\ \hline
 \multicolumn{8}{l}{} \\
 \multicolumn{8}{l}{\textbf{Planetary-like Environments}} \\
 \hline 
 MADMAX \cite{madmax} & Hand-held & Desert & \cmark(M) & \cmark(M) & \cmark & \xmark & D-GNSS \\
 S3LI-Etna \cite{giubilato2022challenges} & Hand-held & Volcano & \cmark(M) & \cmark(M) & \cmark & \cmark(SS) & D-GNSS \\
 Canadian Planetary Emulation \cite{canadian} & Rover & Planet. testbed & \xmark & \xmark & \xmark & \cmark(M) & D-GNSS \\
 TAIL-Plus \cite{wang2024tailplus} & Quadruped & Sandy beach & \cmark(C) & \cmark(C) & \cmark & \cmark(M) & D-GNSS \\
 Erfoud \cite{lacroix2020erfoud} & Rover & Desert & \cmark(M) & \cmark(M) & \cmark & \cmark(M) & D-GNSS \\
 Katwijk Beach \cite{hewitt2018katwijk} & Rover & Sandy beach & \cmark(C) & \cmark(C) & \cmark & \cmark(M) & D-GNSS \\
 Mars C-SLAM \cite{lajoie2025multi}  & Multi-rover & Planet. testbed & \xmark & \xmark & \cmark & \cmark(M) & D-GNSS  \\ 
 \hline
 
 \multicolumn{8}{l}{} \\
 \multicolumn{8}{l}{\textbf{Unstructured Natural Terrestrial Environments}} \\ \hline
 ROVER \cite{schmidt2025rover}  & Rover & Gardens & \cmark(C) & \cmark(C) & \cmark & \xmark & Prism \\ 
 WildPlaces \cite{knights2023wild}  & Hand-held & Forest & \cmark(C) & \cmark(C) & \cmark & \cmark(M) & SLAM \\ 
 Botanic Garden \cite{liu2024botanicgarden}  & Rover & Gardens & \cmark(C) & \cmark(C) & \cmark & \cmark(M+SS) & SLAM \\ 
 DiTer++ \cite{kim2024diter++}  & Quadruped & Gardens & \cmark(C) & \xmark & \cmark & \cmark(M) & SLAM \\ 
 GreenBot \cite{canadas2024greenbot}  & Rover & Forest & \cmark(C) & \cmark(C) & \cmark & \cmark(M) & D-GNSS \\ 
 Great Outdoors \cite{jiang2025go} &  Rover & Forest & \cmark(C) & \cmark(C) & \cmark & \cmark(M) & D-GNSS \\ 
 \hline 
 
 \multicolumn{8}{l}{} \\
 \multicolumn{8}{l}{\textbf{Artificial Structured Environments}} \\ \hline
 IndoorMCD \cite{10016760} & Platform & Households & \cmark(C) & \cmark(C) & \cmark & \xmark & Tracker \\
 M2DGR \cite{yin2021m2dgr}  & Rover & Indoor & \cmark(C) & \cmark(C) & \cmark & \cmark(M) & Multiple \\ 
 S3E \cite{feng2024s3e}  & Rover & Uni. Campus & \cmark(C) & \cmark(C) & \cmark & \cmark(M) & D-GNSS \\ 
 Multi-Floor \cite{kaveti2023challenges} & Rover & Indoor & \cmark(C) & \cmark(C) & \cmark & \cmark(M) & SLAM \\ 
 \hline
 
 \multicolumn{8}{l}{} \\
 \multicolumn{8}{l}{\textbf{Other}} \\ \hline
 SubT-MRS \cite{zhao2024subtmrs}  & Multiple & Multiple & \cmark(C) & \cmark(C) & \cmark & \cmark(M+SS) & Multiple \\ 
 \hline

 \multicolumn{8}{l}{} \\ \hline
 \rowcolor[gray]{.9}
 \textbf{S3LI-Vulcano} & Hand-held & Multiple & \cmark(C) & \cmark(C) & \cmark & \cmark(SS) & D-GNSS \\
 \hline
\end{tabular}
\end{threeparttable}
\end{table*}

To foster research in the fields of localization and mapping in planetary settings, and provide data to researchers to develop, test and tune algorithms for this purpose, we release a multi-modal dataset recorded from an hand-held device, that mimicks the height and point of view of a mobile robotic explorer. The sensor setup, named \textbf{S3LI} (\textbf{S}tereo, \textbf{S}olid-\textbf{S}tate \textbf{L}iDAR \textbf{I}nertial) comprises a pair of RGB cameras in horizontal stereo configuration, a compact LiDAR with a MEMS-actuated reflective mirror, an industrial-grade IMU and a GNSS antenna. All sensors are carefully calibrated on the field, and time-synchronized via Precision Time Protocol (PTP). A differential GNSS solution is also provided for evaluation and global referencing. Contrarily to existing multi-modal datasets for SLAM, our sensor setup is intended to explore future possibilities for perception systems suitable for planetary exploration thanks to traditional camera sensors and an alternative LiDAR technology to common spinning mirrors. 

The dataset is recorded in the island of Vulcano, Eolian Islands, Sicily, characterized by peculiar geological properties resulting from the fusion of several volcanoes, of which one currently active. The island presents a large variety of natural environments with appearance and structures that resemble settings for future scientific exploration missions. Several sequences observe different traversable terrains and geological features spanning from lava paths or sedimentary structures, posing different challenges towards localization accuracy and long-term consistency of mapping pipelines, offering furthermore various options towards efficient fusion of complementary modalities. In addition to this, we release sequences that observe basaltic lava structures, water and vegetation, introducing dynamic elements that disturb traditional algorithmic pipelines.

This paper comprises a collection of Robot Operating System (ROS) recordings of raw sensor data, including calibration sequences. Accompanying the paper, we release a complementary toolkit as an open-source tool to replicate calibration procedures and post-processing of the data, e.g., generating differential GNSS solutions. We furthermore release example scripts and configuration files to execute existing state of the art SLAM algorithms as an aid towards the utilization of the data. 

\section{Related Works}

Several datasets have been released over the years to address the development and testing of SLAM performances. In literature, the predominant case for visual- or LiDAR-based SLAM algorithms relates to indoor or artificial scenes \cite{sturm2011towards}, with references to applications such as Augmented Reality, or robotics for constructions \cite{helmberger2022hilti}, targeting often short sequences in very controlled conditions \cite{schops2019bad}. A large body of literature relates otherwise to the case of autonomous driving \cite{geiger2013vision,barnes2020oxford}, presenting long sequences from the perspective of a moving car, traversing static or dynamic scenarios and across diverse weather and lighting characteristics \cite{wenzel20204seasons}. This large amount of data, however, fails to address specific conditions which are typical of what robots encounter in the field. First, road-bound vehicles can traverse environments in two directions, imposing significant constraints on the diversity of viewpoints and perspectives and affecting positively the performances of place recognition algorithms for detection of loop closures \cite{lowry2015visual}. Secondly, artificial outdoor environments, such as urban settings, introduce unique elements, both from a visual as well as structural perspective, that facilitate localization. To address the case of a robot rowing in the field, several datasets have been proposed, of which a selected sample is here discussed. 

\subsection{Unstructured and Natural Terrestrial Environments}
A relatively large body of literature exists that relates to natural unstructured terrestrial environments. Several datasets address the case of mapping in forest scenes, with target use-cases of either survey of trees and forest digitalization or large-scale navigation from wheeled vehicles. Wild-Places \cite{knights2023wild} addresses the task of LiDAR-based place recognition in kilometer-long sequences. The authors provide LiDAR submaps, accompanied by RGB imagery, with significant overlap, to investigate loop closure detection or re-localization in the context of multi-session mapping. Similarly, Great Outdoors \cite{jiang2025go} present data captured in forests, this time from the perspective of a rover system, and adding semantic annotation to RGB data. Botanic Garden \cite{liu2024botanicgarden}, instead, addresses the case of under-canopy localization and mapping in the specific context of botanical gardens, providing precise time-synchronization and calibration between RGB and multiple LiDAR sensors, as well as semantic annotations. More variety, in the context of motion patterns, as well as weather and lighting conditions, is introduced by the ROVER \cite{schmidt2025rover} dataset, capturing sequences in parks and gardens recorded from the perspective of a robotic platform. While exposing significant challenges with respect to aliasing and ambiguity of appearance, these datasets often present regularities introduced by driving along paths or unpaved roads, therefore facilitating relocalization tasks. Furthermore, the abundance of vegetation makes these data not relevant for benchmarking of algorithms targeted at planetary-like environments

\subsection{Planetary-Like Environments}
Several attempts at gathering data which is representative of planetary environments have been made, generally targeting locations which, for their geological properties and overall appearance, resemble parts of the lunar or martial soil. The MADMAX \cite{madmax} dataset, collected in three locations in the Moroccan desert, provide images and depth from a stereo setup, with accurate D-GNSS ground truth and orientations. While providing diverse sequences of particular motion patterns, aimed at traversing or mapping tasks, the presence of a single sensing modality limits the applicability to uni-modal settings. Collected in a geographically close location, the Erfoud \cite{lacroix2020erfoud} dataset provides, in addition to visual and multi-spectral data, LiDAR scans from two rovers, but is limited in variety of environments and scenes. Collected in artificial planetary analogous environments, the Canadian Planetary Emulation dataset \cite{canadian} as well as the Katwijk beach dataset \cite{hewitt2018katwijk} provide rover sequences with stereo, or multi-camera, and LiDAR sequences or fully or partially artificial settings, aiming at replicating natural rocky structures that can be observed in martian scenes. Similary to the Erfoud dataset, however, the variety of scenes is limited to the circumscribed test environment and, true or artificial, geological features. The TAIL dataset \cite{wang2024tailplus} targets quadruped robots in beach settings, provides multi-modal inputs and revisit of scenes across diverse daytimes, but limited in the scene diversity.

\section{The S3Li Vulcano Dataset}

\begin{table*}[htp]
\caption{Overview of sequences from the S3LI Vulcano Dataset}
\begin{threeparttable}
  \centering
  \setlength{\tabcolsep}{1pt}
  \renewcommand\tabularxcolumn[1]{>{\centering\arraybackslash}m{#1}}
  \renewcommand{\arraystretch}{1.2}
  \newcolumntype{L}{>{\raggedright\arraybackslash}X}
  \newcolumntype{C}{>{\centering\arraybackslash}X}
  \newcolumntype{R}{>{\raggedleft\arraybackslash}X}
  \begin{tabularx}{\textwidth}{|c|X|X|c|L|}
    \hline
    \textbf{Seq. Name} & 
    \textbf{Example Image} & 
    \textbf{GPS Track} \tnote{a} & 
    \textbf{Duration} & 
    \textbf{Notes} \\
    \hline\hline
    \cmtt{waterfront} & 
    \includegraphics[width=.2\textwidth]{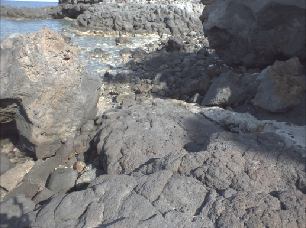} &
    \vspace{5pt} 
    \includegraphics[width=.23\textwidth]{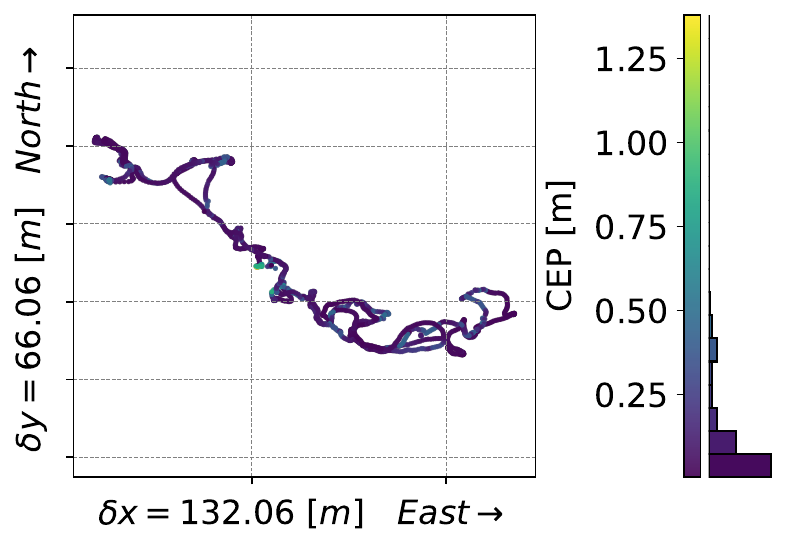} & 
    24:47s
    & 
    Long sequence on a waterfront of basaltic rocks and vegetation. Plenty of opportunities for re-localization, and exploitation of multi-modality. 
    \\
    \hline
    \cmtt{vegetation} &
    \includegraphics[width=.2\textwidth]{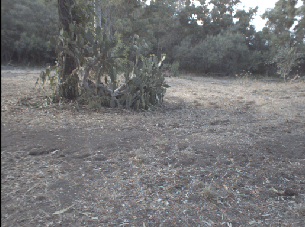} &
    \vspace{5pt} \includegraphics[width=.23\textwidth]{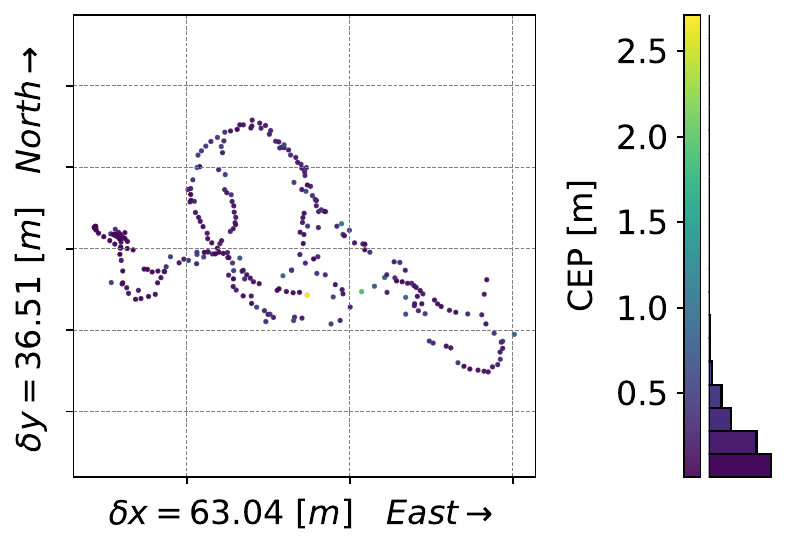} & 
    3:59s
    & 
    Simple sequence with loops under tree canopies and dry grass. Opportunities for structure-based loop closure detection, e.g. tree trunk detection and mapping. 
    \\ \hline
    \cmtt{capo\_grillo} & 
    \includegraphics[width=.2\textwidth]{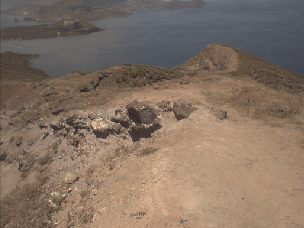} &
    \vspace{5pt} \includegraphics[width=.23\textwidth]{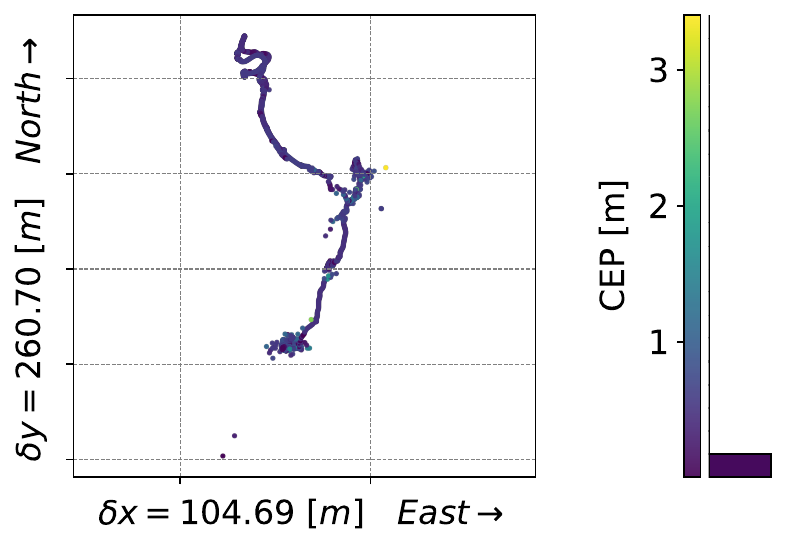} & 
    27:36s 
    & 
    Long sequence on a hiking path, covering diverse environments, from grassy to rocky. Some opportunities for loop closure but incomplete GNSS track. 
    \\ \hline
    \cmtt{capo\_grillo\_2} &
    \includegraphics[width=.2\textwidth]{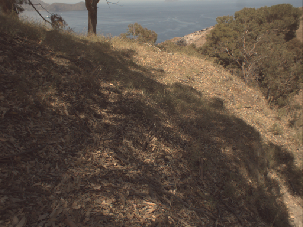} &
    \vspace{5pt} \includegraphics[width=.23\textwidth]{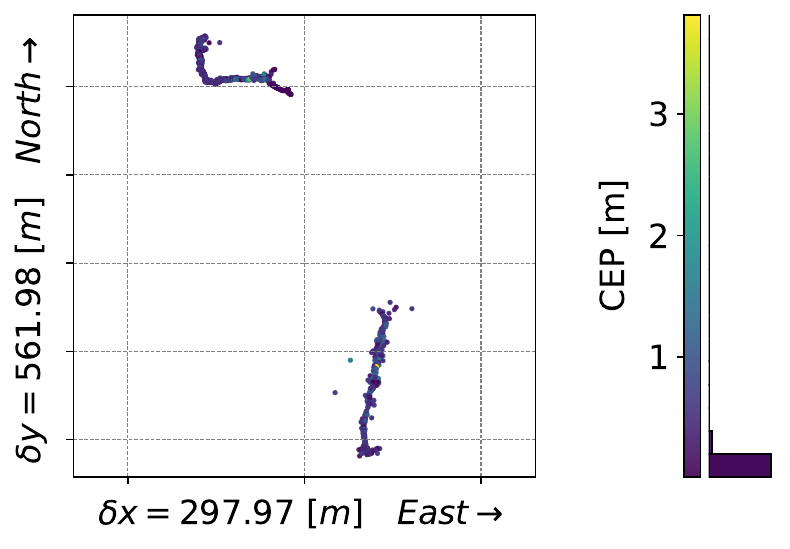} & 
    18:40s
    & 
    Shorter sequence with similar character to \cmtt{capo\_grillo} focusing more on grassy areas but with far-range structures to investigate multi-modal mapping
    \\ \hline
    \cmtt{straight\_path} & 
    \includegraphics[width=.2\textwidth]{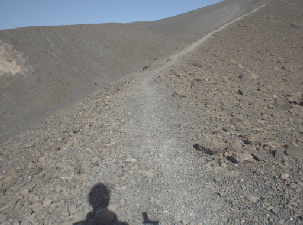} &
    \vspace{5pt}
    \includegraphics[width=.23\textwidth]{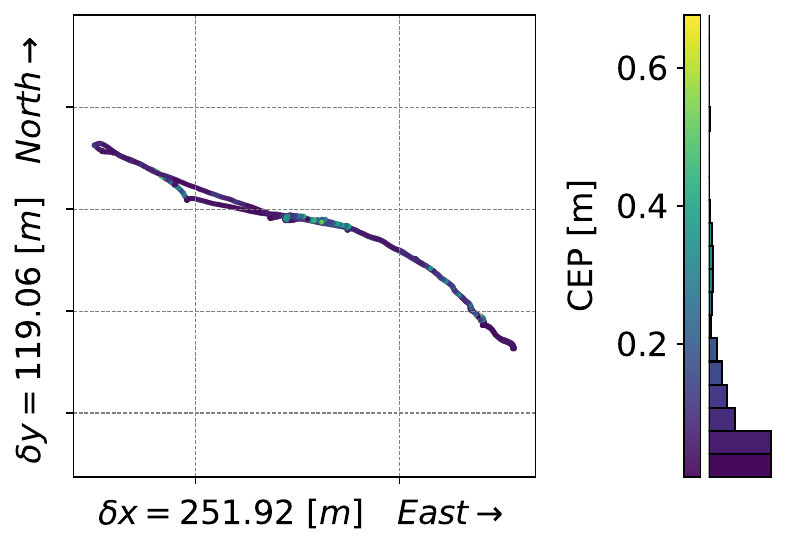} & 
    12:21s
    & 
    Simple hiking sequence on a straight path with repetitive visual and structural features. Useful to investigate loop closure from opposing views. 
    \\ \hline
    \cmtt{lava\_tunnel} &
    \includegraphics[width=.2\textwidth]{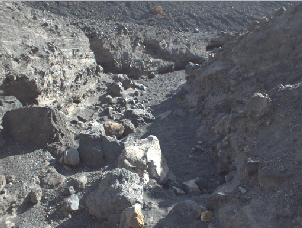} &
    \vspace{5pt} \includegraphics[width=.23\textwidth]{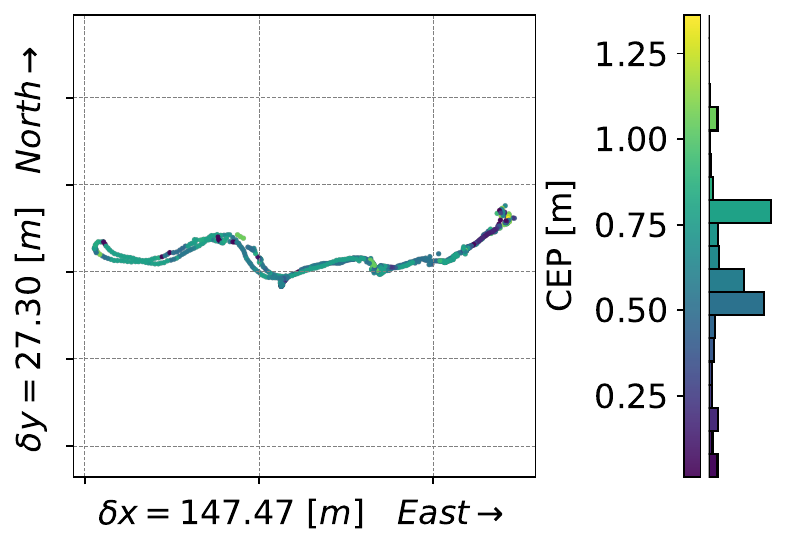} & 
    8:47s
    & 
    Unique sequence on an old lava tunnel, with walls up to 2 meters high. Offers opportunities to study loop closure techniques based on natural structures
    \\ \hline
    \cmtt{moon\_lake} &
    \includegraphics[width=.2\textwidth]{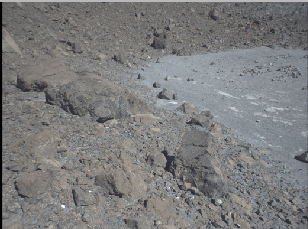} &
    \vspace{5pt} \includegraphics[width=.23\textwidth]{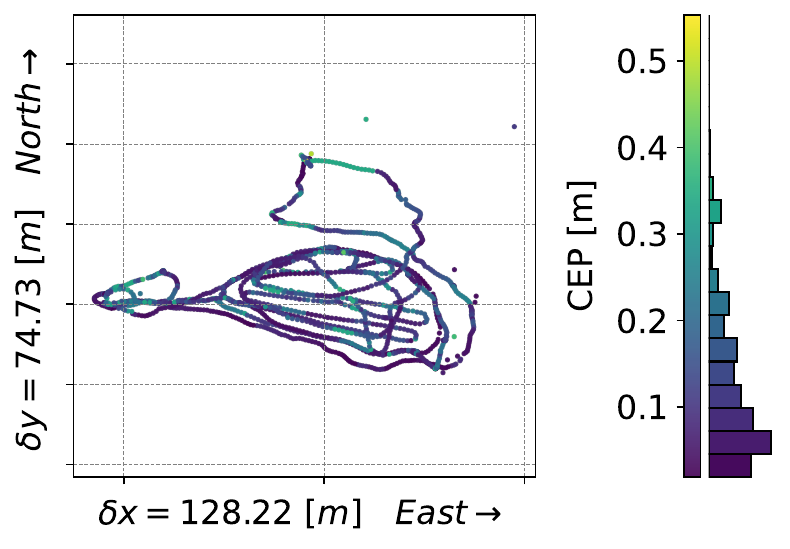} & 
    22:28s
    & 
        Long sequence walking in a flat "lake"-like basin of gray sand, inside a small crated of orange rocks. Numerous revisits and repeated traverses in alternating directions to evaluate loop closure and mapping.

    \\ \hline
  \end{tabularx}
  \begin{tablenotes}
        \item[a] The trajectories plotted are filtered with respect to the \textit{quality} metric, and plotted only for $\mathbf{Q>=2}$ (floating solution or RTK Fix)
    \end{tablenotes}
\end{threeparttable}
\label{tab::sequences}
\end{table*}

The S3LI "Vulcano" dataset is an extension of the S3LI dataset \cite{giubilato2022challenges}, recorded on a plateau near the summit of Mt. Etna, Sicily, and around the Cisternazza crater. The original S3LI dataset targeted a "moon-like" environment for its appearance, characterized by black and dark gray slopes of lava ash, with ambiguous terrain features and lack of outstanding structures. The overarching idea of both iterations is to provide multi-modal data, where a LiDAR modality is recorded from the perspective of a Solid-State LiDAR system, with a MEMS-actuated oscillating mirror, opposed to traditional spinning mirror LiDAR technologies. This, in conjunction with traditional visual and inertial sensing, drives the development of state estimation and SLAM algorithms, as well as place recognition for detection of loop closures, specifically in the context of potential space-qualifiable sensing technologies.

The S3LI Vulcano dataset provides complementary data to the original dataset improving the variety of data inputs and scenes. More specifically: \begin{itemize}
    \item we provide \textbf{RGB} \textbf{data} instead of monochrome data, which is particularly important in the context of \textbf{semantic segmentation} and classification of objects and terrains \cite{muller2023uncertainty}.
    \item the LiDAR modality exploits a second iteration of the original sensor, which increases the \textbf{minimum sensing range} to 1.5 meters, better overlapping the range of operation of depth from stereo.
    \item we record data that include \textbf{sand}, \textbf{rocks}, \textbf{water} and \textbf{vegetation}, driving towards a better balance between terrain classes and features.
\end{itemize}

Table~\ref{tab::sequences} provides an organized overview of the seven sequences that the dataset provides. The \cmtt{waterfront} sequence is recorded on a bay where dark basaltic rock creates a high promontory. In this sequence, dry grass mixes with rich brown terrains and dark rock, and frequent revisits of the same places allows to test place recognition algorithms across different viewpoints, as well as consistency of 3D mapping techniques. The \cmtt{vegetation}, \cmtt{capo\_grillo} and \cmtt{capo\_grillo\_2} observe a mixture of vegetation and dark soil, often looking at the terrain from various distances, where multi-modality can be well exploited. The sequences \cmtt{straight\_path}, \cmtt{lava\_tunnel} and \cmtt{moon\_lake} are all recorded near the summit of \textit{Gran Cratere}, the summit of the volcano that gives the island its name. Here, complex volcanic features, such as an old tunnel from lava flow, or the basin of an old \textit{caldera} with a constellation of scattered rocks and boulders, gives many options to the user to exploit different terrain geometries. 

\subsection{Sensor Setup}
The sensor setup, visible in Figure~\ref{fig:s3li_setup} and summarized in Table~\ref{tab:s3li_setup}, comprises:
\begin{itemize}
    \item a pair of hardware-triggered AVT Manta G-319C RGB cameras, capturing images at a fixed frame rate of 10 Hz
    \item a solid-state Blickfeld Cube-1 LiDAR, capturing scans at a rate of ~4.7 Hz with a Field of View of 70 degrees horizontal and 30 degrees vertical, for a total of 17800 points per scan
    \item an XSens MTi-G 10 IMU connected via USB to the main pc
    \item a GNSS antenna connected to an UBlox module, and recording raw GNSS binary dumps to disk, for later offline processing of a differential solution
\end{itemize}

Sensors are time-synchronized with the clock of the main PC, an Intel NUC equipped with an i7 processor, using the PTP4l protocol, supported by the cameras and the LiDAR. IMU measurements are instead assigned timestamps at the time of arrival into the NUC and a value for the latency is later estimated within a Visual-Inertial Odometry framework. All sensor drivers, except for the UBlox device, are interfaced with the Robot Operating System (ROS) and the data is recorded and packaged using the \textit{rosbag} serialization format. 

\begin{table}[tp]
    \centering
    \renewcommand{\arraystretch}{1.5}
    \setlength{\tabcolsep}{10pt}
    \caption{List of components of the S3LI setup}
    \small
    \setcellgapes{2pt} 
    \makegapedcells
    \begin{tabular}{|l|c|l|}
        \hline
         \textbf{Component} & \textbf{Rate} & \textbf{Characteristics} \\ \hline\hline
         AVT G-319C (2x) & 10 Hz & 
        \makecell[l]{ 
            \tabitem 2 MPx images \\
             \tabitem Color images \\
             \tabitem Hardware triggered}
         \\ \hline
         XSens MTi-G 10 & 400 Hz & - \\ \hline
         Blickfeld Cube 1 & 4.7 Hz & 
         \makecell[l]{
         \tabitem Solid State LiDAR \\
         \tabitem 70(H)x30(V) FoV \\
         \tabitem Elliptical pattern} \\ \hline
         UBlox GNSS & 5 Hz & 
         \makecell[l]{\tabitem One GNSS Antenna \\ 
         \tabitem Offline D-GNSS} \\ \hline
         Intel NUC & - & \makecell[l]{
         \tabitem Intel i7 CPU \\
         \tabitem 16 Gb RAM } \\
         \hline
    \end{tabular}
    \label{tab:s3li_setup}
\end{table}

\subsection{Calibration and Post-Processing}

\subsubsection{Camera Calibration} 
Cameras are calibrated as a stereo setup using DLR's CalDe / Callab \footnote{\url{https://www.dlr.de/en/rm/research/publications-and-downloads/software}} \cite{callab}. Special calibration board with reference markers allow the camera to observe partial views of the checkerboard, with the important benefit for stereo camera setups to cover the entire image with corner points regardless of the overlap between views. 


\subsubsection{IMU-Camera Calibration}
IMU-to-camera calibration is performed using Kalibr \cite{rehder2016extending}, an open-source multi-camera-IMU calibration toolkit. Kalibr uses custom calibration patterns relying on arrays of AprilTag markers and determines 6D transformations between sensors, as well as clock differences. We provide the best multi-camera-IMU calibration results obtained with respect to the original unprocessed data, therefore full resolution images as in Table~\ref{tab:s3li_setup}.

In practise, however, the calibration from the Kalibr toolkit provides an initial approximation of the extrinsic parameters to either R-VIO2 \cite{huai2022square,huai2024consistent} or VINS-Fusion \cite{qin2018vins,qin2019general}, state of the art visual-inertial odometry algorithms that have the ability to estimate online, i.e., during motion, 1) the transformation from the IMU frame to the left camera frame, and 2) the time offset between the IMU timestamps and the left camera timestamps, assuming that there is time-varying latency in the loop. We experience that, when evaluating the performances of these algorithms, online optimization of the extrinsic parameters from cameras to IMU provide generally the best results and robustness.

\subsection{Generation of Differential GNSS Solution for Ground Truth}
Although a GNSS track was recorded from the S3LI setup with a single antenna, we provide a differential solution generated offline using the RTKLIB\footnote{\url{rtklib.com}}. GNSS data from the UBlox module was recorded as a sequence of NMEA sentences. These include all the necessary information to compute a differential solution with a second reference static antenna on the ground. As the base antenna, the closest one to the Vulcano island and part of the EUREF Permanent GNSS Network is the one in Noto, Sicily. After converting data to the RINEX format, a differential solution was computed for all sequences and saved in the \cmtt{.pos} format. Details on how to use, obtain or generate data are included in the dataset repository. Table~\ref{tab::sequences} reports the D-GNSS tracks for all sequences, including points with quality level Q$\leq$2, where 1 is full fix (10$^{-2}$ to 10$^{-1}$ meters accuracy) and 2 is floating solution (10$^{-1}$ to 10$^0$ meters accuracy). Each D-GNSS track is color-coded with a Circular Error Projection (CEP) estimate based on the standard deviations of the position solution in the directions of latitude and longitude.

\section{Example Applications}
\label{sec:eval}
\begin{table}[t]
\begin{threeparttable}
    \centering
    \setlength{\tabcolsep}{3pt}
    \renewcommand{\arraystretch}{1.5}
    \setcellgapes{2pt} 
    \makegapedcells
    \caption[Performance comparison of V-SLAM algorithms]{
    Performance comparison of a selected sample of V-SLAM algorithms.
    RMSE is expressed in meters, as well as relative to the length of the ground-truth trajectory.
    }
    \footnotesize
    \begin{tabular}{|l|c|c|c|c|c|c|c|}
        \hline
         \small \textbf{Name} & \multicolumn{7}{|c|}{\small \textbf{RMSE [m]}} \\ \hline
         \hline
         \textbf{R-VIO2} & 
         \makecell{2.29 \\  0.28\%} & 
         N/I &  
         \makecell{0.56 \\ 0.03\%} &  
         \makecell{0.39 \\ 0.04\%} & 
         \makecell{0.92 \\  0.14\%} & 
         \makecell{1.02 \\  0.14\%} &
         \makecell{1.65 \\  0.12\%} \\ \hline
         \textbf{VINS-Fusion\tnote{a}} &
         \makecell{35.5 \\  8.73\%} & 
         \makecell{2.07 \\  0.28\%} &  
         F &  
         F & 
         N/I &
         N/I & 
         \makecell{1.62 \\ \footnotesize 0.13\%} \\ \hline
         \textbf{SqrtVINS} &
         N/I & 
         N/I &  
         N/I &  
         N/I & 
         N/I &
         \makecell{1.04 \\  0.15\%} & 
         N/I \\ \hline
         &  \rotatebox{90}{\cmtt{waterfront}} &  \rotatebox{90}{\cmtt{vegetation}} & 
          \rotatebox{90}{\cmtt{capo\_grillo\tnote{b} \ }} &  \rotatebox{90}{\cmtt{capo\_grillo\_2\tnote{b} \ }} &
          \rotatebox{90}{\cmtt{straight\_path}} &  \rotatebox{90}{\cmtt{lava\_tunnel \ }} & 
          \rotatebox{90}{\cmtt{moon\_lake}} \\ \hline
    \end{tabular}
    \label{tab:slam_eval}\begin{tablenotes}
        \item[a] evaluated using the \textit{global\_fusion} mode, i.e., with loop closure.
        \item[b] RMSE is evaluated using a RANSAC implementation of the Horn's algorithm, due to the significant quantity of noisy D-GNSS data points, see Tab.~\ref{tab::sequences}
    \end{tablenotes}
\end{threeparttable}
\end{table}

\begin{figure*}[tp]
\centering
\begin{tikzpicture}[every node/.style={anchor=west}]

  \node (img) at (0,0) {\includegraphics[width=.95\linewidth]{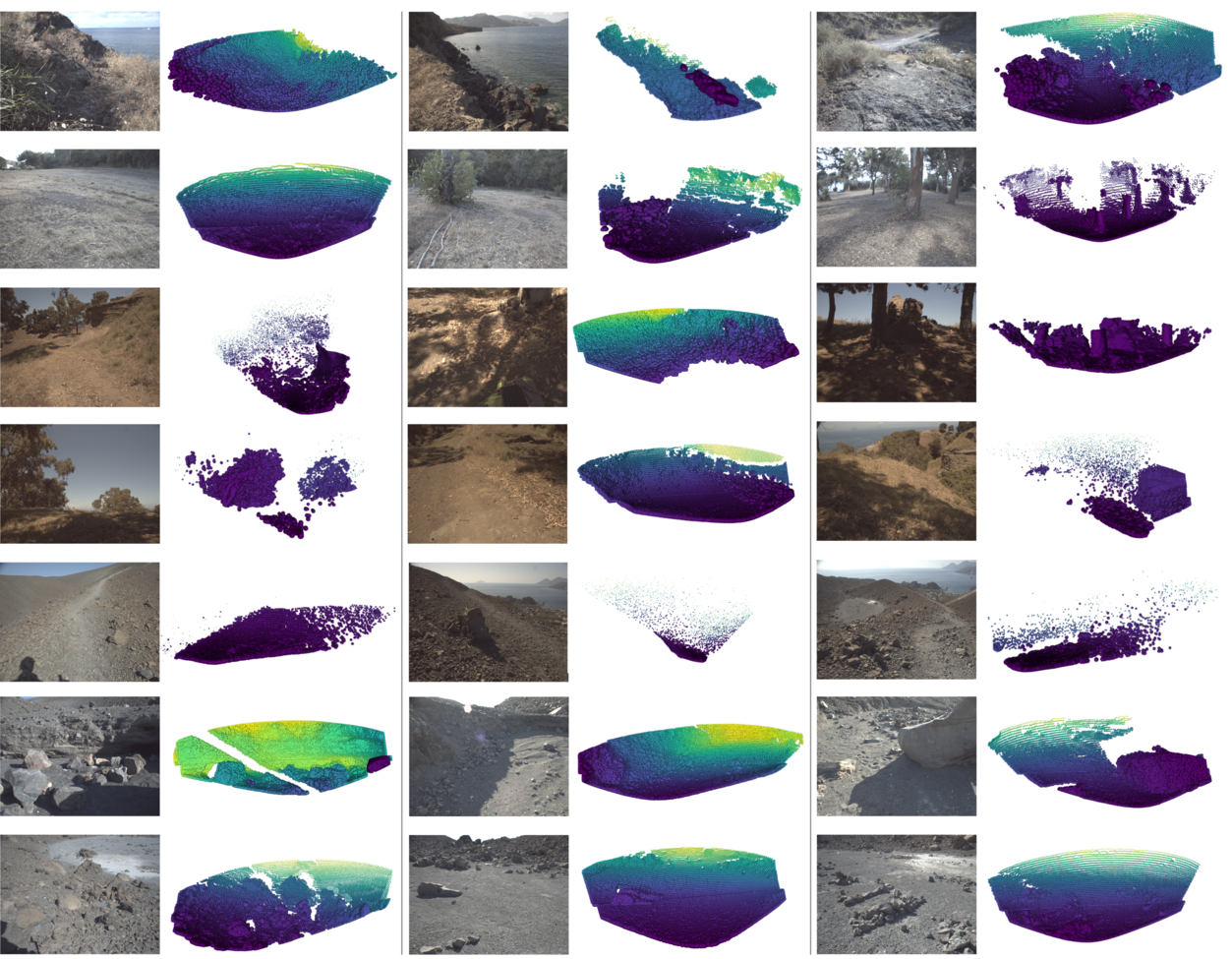}};
  \pgfmathsetmacro{\rows}{7}
  \pgfmathsetmacro{\imgheight}{8.25} 
  \foreach \i/\labeltext [evaluate=\i as \y using {\imgheight - (\i)*(1.85)}] in {
  1/\cmtt{\small waterfr.},
  2/\cmtt{\small vegetat.},
  3/\cmtt{\small c.\_grillo},
  4/\cmtt{\small c.\_grillo\_2},
  5/\cmtt{\small st.\_path},
  6/\cmtt{\small lava\_tun.},
  7/\cmtt{\small moon\_lake}
} {
  \node[anchor=east, rotate=90] at ($(img.west) + (-0.3cm, \y cm)$) {\labeltext};
}
  
\end{tikzpicture}
\caption{Graphical examples of associated RGB images and LiDAR scans from the dataset sequences. For each sequence, 3 samples of corresponding images and scans are provided}
\label{fig:example_lidar_scans}
\end{figure*}

Compared to traditional datasets for benchmarking and evaluation of visual \cite{fontan2025vslam} or multi-modal SLAM techniques, where the scenes are often short in trajectory length and duration, and the scene is usually set in a controlled manner, operating rovers to achieve scientific or infrastructural inspection tasks often implies the absence of bounds on trajectory length and map size. The S3LI Vulcano dataset, in addition to the previous iteration on Mt. Etna, provides sequences with strong resemblance to actual traversing or large-scale mapping operation of entirely unstructured natural environments. Traditional \textit{full} visual-SLAM techniques, such as ORB-SLAM \cite{campos2021orb}, which require maintenance of feature maps without particular partitioning or marginalization strategies, often fail to guarantee asymptotic performance metrics that are appropriate for real-time operation requirements. In Table~\ref{tab:slam_eval} is a performance evaluation given as an example application case for visual, or visual-inertial SLAM techniques. The real-time camera pose estimates are aligned using Horn's algorithm, fixing the initial position and determining only the rotation for the best alignment. The total RMSE (Root Mean Square Error) for all the position residuals, as well as the RMSE normalized on the trajectory length, provide an insight on how well an algorithm fits the real camera movement, although it has to be considered that large trajectory errors for odometry measurements can depends on single failure events along the path, and are not indicative of constant drift. 

Table \ref{tab:slam_eval} reports an simple evaluation of Visual-Inertial SLAM algorithms in various configurations. 
For these tests, RGB images are downsampled by a factor 4 to facilitate real-time performances in stereo rectification and depth estimation. The first, R-VIO2 \cite{huai2024consistent} is a monocular Visual-Inertial Odometry (VIO) algorithm. It uses a single camera image stream and IMU measurements, and requires static initialization, therefore the algorithm must be started without exciting the IMU axes in order to provide a first bias estimation. The second, VINS-Fusion \cite{qin2019general}, is a VIO algorithm, used in stereo configuration, with a pose-graph optimization stage, that performs appearance-based loop closure detection and place recognition with BRIEF features to correct the accumulated drift. Errors are expressed as the Root Mean Square Errors between estimated camera poses and corresponding ground truth positions, after aligning the trajectories using the Horn's quaternion method with a fixed null translation. 
The table reports with the nomenclature "N/I" the case where the algorithm fails to initialize and therefore does not provide any pose output. Otherwise, if the algorithm fails along the trajectory, estimating less than 30\% of the camera poses, the table denotes the result as "F". As expected, providing unstructured natural scenes rich local information, pose estimation is a task that can be successfully accomplished by modern algorithms, while the challenges lays on relocalization due to their ambiguity. 

\subsection{Multi-modal Place Recognition}
\begin{figure*}[tp]
    \centering
    \includegraphics[width=\textwidth]{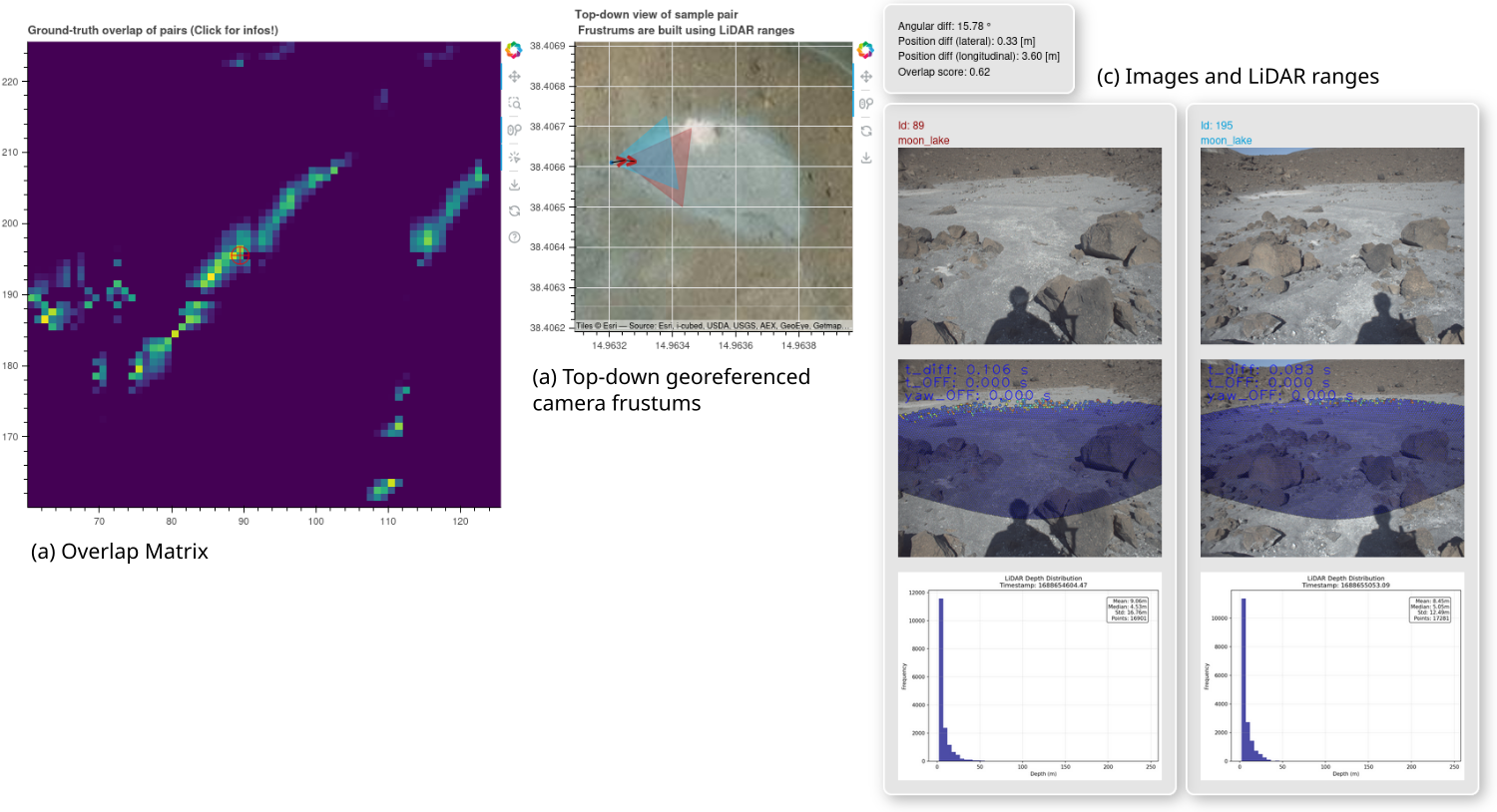}
    \caption{Example View from the place recognition toolkit. On the left, a clickable overlap matrix shows, with colors from blue to yellow, the amount of overlap between two views. When clicked, the corresponding position and FoV (Field of View) of the two images is shown on the map in the center, while images, pointcloud overlaps and point depth histograms are shown on the right.}
    \label{fig:overlap_toolkit}
\end{figure*}

In order to foster research on multi-modal place recognition, we provide as part of the toolkit, functionalities to generate geo-localised pairs of images and LiDAR scans with 6D ground reference (position w.r.t. an arbitrary origin, and orientation w.r.t. an ENU reference frame). These can be used to train, fine-tune and test models for uni- or multi-modal place recognition. That is, their capabilities to recall previously visited scenes. The user can then implement its own custom mechanism to assign ground truth labels, given the global localization information, as well as depth distribution statistics for the pointclouds. Figure \ref{fig:overlap_toolkit} shows the graphical output of the toolkit, allowing the user to carefully inspect the generated data and 6D ground truth poses of data point pairs. In the background, the pipeline stores all information as \textit{pickle} files, to be loaded later in a second time from the user for place recognition tasks. 

\section{Conclusions}
With this paper we release \textbf{S3LI Vulcano}, an extension to the original S3LI dataset \cite{giubilato2022challenges}. The dataset presents additional data, recorded in a selection of severely unstructured natural environments, presenting basaltic rocks, sand, vegetation and water, from the Vulcano island, part of the Aeolian islands in Sicily, Italy. The dataset contains RGB images, pointclouds from an solid-state LiDAR, inertial and GNSS data, in form of ROS bagfiles. We demonstrate an example use case as for the development and testing of visual or multi-modal SLAM algorithms. Finally, we release a software toolkit that generates samples with global ground truth to train and test multi-modal place recognition algorithms. 

\acknowledgements 
This work was supported by the Helmholtz Association project iFOODis (contract number KA2-HSC-06).

\bibliographystyle{IEEEtran}
\bibliography{refs}

\thebiography
\begin{biographywithpic}
{Riccardo Giubilato}{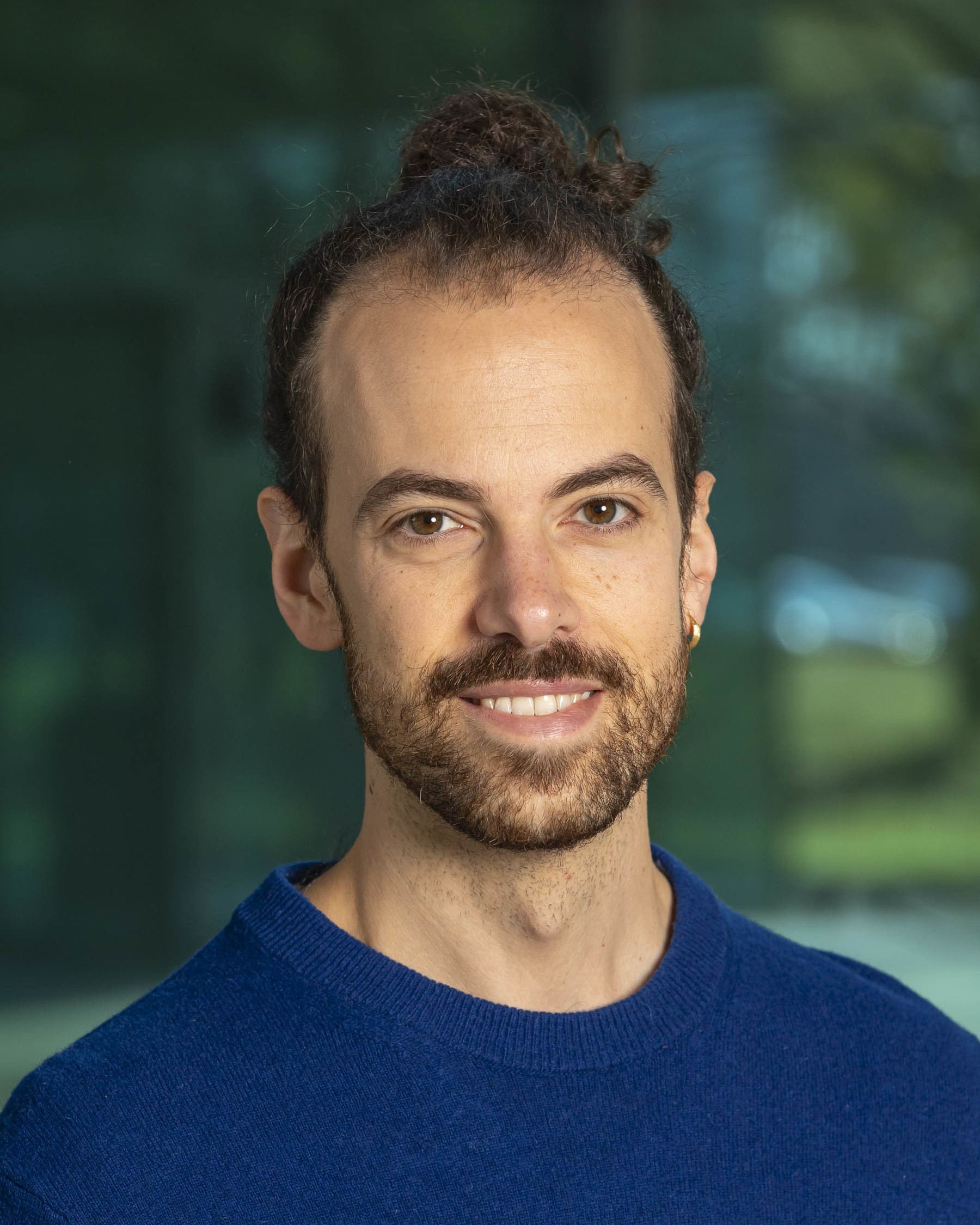}
is a researcher at the Institute of Robotics and Mechatronics, German Aerospace Center (DLR), in the Dept. of Perception and Cognition since 2019. He leads the Intelligent Exploration and Mapping group, focusing on development of localization, semantic and long-term mapping, and multi-agent exploration approaches to facilitate high-level understanding and exploitation of known or unknown environments. 
He holds a Ph.D. in Space Sciences, Technologies and Measurements from CISAS "Giuseppe Colombo", and Master's and Bachelor's degree in Aerospace Engineering from the University of Padova, Italy.  
\end{biographywithpic} 

\begin{biographywithpic}
{Marcus Gerhard Müller}{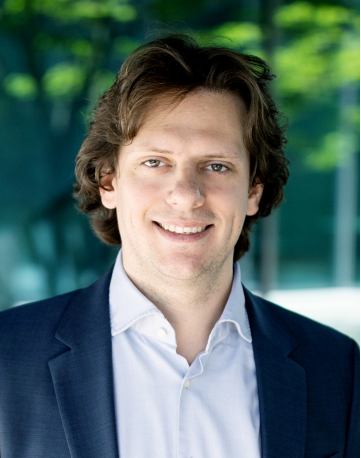}
is a researcher in the
department of ”Perception and Cognition” at the German Aerospace Center
(DLR) since 2016 and Ph.D. student at
ETH Zurich. He is the leader of the
MAV Exploration Team at the Institute of
Robotics and Mechatronics (DLR-RM),
where he is working on autonomous navigation algorithms for MAVs. Before
joining DLR he conducted research at
the Jet Propulsion Laboratory (JPL) of NASA in Pasadena,
USA, where he worked on visual inertial navigation for
MAVs and on radar signal processing. Marcus received his
Master’s and Bachelor’s degree in Electrical Engineering
from the University of Siegen, Germany
\end{biographywithpic} 

\begin{biographywithpic}
{Marco Sewtz}{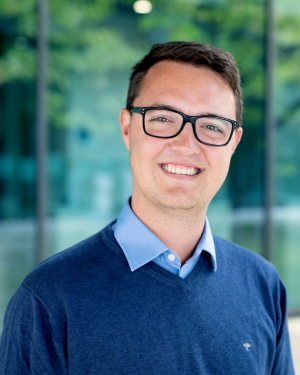}
is the lead for software
and interfaces of the new lunar exploration rover at the German Aerospace
Center (DLR) and responsible for the development of AI technology on robotic
space systems. He received his B.Eng.
at the University of Applied Sciences of
Munich, his M.Sc. at the Technical University of Munich (TUM) and his PhD
at the Karlsruhe Institute of Technology
(KIT). Before his current role, he worked as an electrical
designer for high-performance processing modules for space
hardware at Airbus Defence and Space
\end{biographywithpic} 

\begin{biographywithpic}
{Laura Alejandra Encinar Gonzalez}{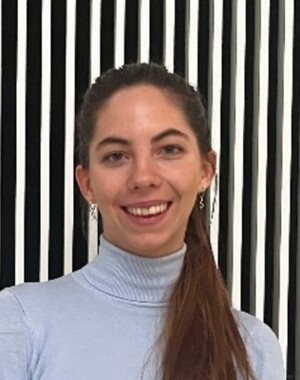}
Laura Alejandra Encinar Gonzalez received her BSc in Mathematics and Computer Science from UPM in 2023 and her MSc in ICT Innovation (Autonomous Systems, AI focus) from Aalto University and KTH in 2025. She worked as a student at MIT Media Lab, focusing on multi-objective optimization for spatial design, and at the German Aerospace Center (DLR), where she developed multimodal systems for perception and localization. Her background combines AI, computer vision, and robotics, with a strong interest in real-time autonomous systems.
\end{biographywithpic} 

\begin{biographywithpic}
{John Folkesson}{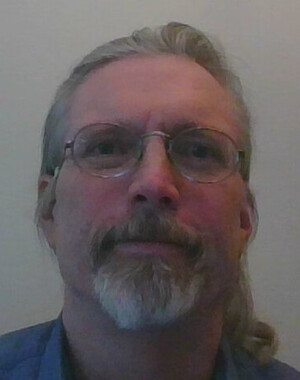}
John Folkesson received his PhD degree from KTH Royal Institute of Technology, Sweden,  in 2006.  He was a postdoc at MIT until 2008 and then was a research Engineer at MIT until 2010.  In 2010 he returned to KTH as a Research Engineer.   He is currently a full Professor of Robotics at KTH. Since returning to KTH in 2010, Dr. John Folkesson has been instrumental in establishing underwater  robotics within the Robotics, Perception, and Learning (RPL) division of the EECS School. As a  Principle Investigator in the Swedish Maritime Robotics Centre (SMaRC), he focuses on  developing planning, perception, and navigation software to enhance autonomy in various  scenarios. 
\end{biographywithpic} 

\begin{biographywithpic}
{Rudolph Triebel}{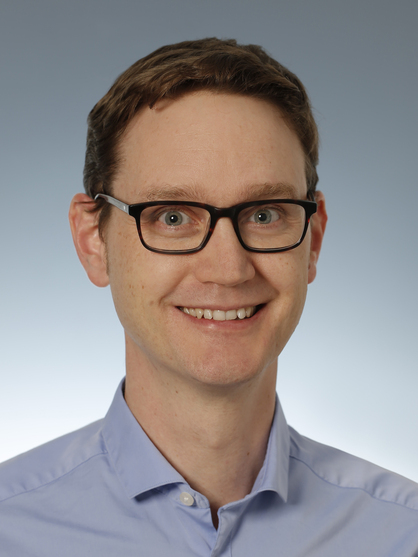}
received his PhD in
2007 from the University of Freiburg
in Germany. From 2007 to 2011, he was a
postdoctoral researcher at ETH Zurich,
working on machine learning algorithms
for robot perception within several EU funded projects. From 2011 to 2013, he
worked in the Mobile Robotics Group
at the University of Oxford, developing unsupervised and
online learning techniques for detection and classification
in mobile robotics and autonomous driving. From 2013 to
2023, Rudolph was a lecturer at the Technical University of
Munich (TUM), teaching master-level courses in Machine
Learning for Computer Vision. In 2015, he became the
leader of the Department of Perception and Cognition at the
Robotics Institute of the German Aerospace Center (DLR),
and in 2023, he was appointed as an university professor at
the Karlsruhe Institute of Technology (KIT) in “Intelligent
Robot Perception”.
\end{biographywithpic}

\end{document}